\documentclass[a4paper,conference]{IEEEtran}

{}
{}
{}

\usepackage{graphicx}
\usepackage{algorithm}
\usepackage{algpseudocode}
\usepackage[usenames, dvipsnames]{color}
\usepackage[table]{xcolor}
\usepackage{amsmath}
\usepackage{amssymb}
\usepackage{enumitem}

\ifCLASSINFOpdf
\else
\fi
  \usepackage[caption=false,font=normalsize,labelfont=sf,textfont=sf]{subfig}

\begin{document}
%
\title{Support Vector Machine (SVM) Recognition Approach adapted to Individual and Touching Moths Counting in Trap Images}


%
\author{\IEEEauthorblockN{M. C. BAKKAY\IEEEauthorrefmark{1},
S. CHAMBON\IEEEauthorrefmark{1},
H. A. RASHWAN\IEEEauthorrefmark{1},
C. LUBAT\IEEEauthorrefmark{2} and
S. BARSOTTI\IEEEauthorrefmark{2}}
\IEEEauthorblockA{\IEEEauthorrefmark{1}University of Toulouse, IRIT, France}
\IEEEauthorblockA{\IEEEauthorrefmark{2}SiConsult, Lab\`{e}ge, France\\
Email: schambon@enseeiht.fr}
}


\maketitle

\begin{abstract}
This paper aims at developing an automatic algorithm for moth recognition from trap images in real-world conditions. This method uses our previous work for detection~\cite{bakkay2017automatic} 
and introduces an adapted classification step. More precisely, SVM classifier is trained with a multi-scale descriptor, Histogram Of Curviness Saliency (HCS). This descriptor is robust to illumination changes and is able to detect and to describe the external and the internal contours of the target insect in multi-scale. The proposed classification method can be trained with a small set of images. Quantitative evaluations show that the proposed method is able to classify insects with higher accuracy (rate of 95.8\%) than the state-of-the art approaches.
\end{abstract}


%
\IEEEpeerreviewmaketitle

\section{Introduction}
\label{sec:introduction}

Insecticides are expensive and dangerous for plants and humans. Therefore, farmers attempt to survey insect species and to evaluate their density in the fields in order to adapt and to reduce the use of insecticides. For this purpose, it is needed to catch insects and then, to manually count these insects in order to analyze the evolution of the insect population and to take the decision for using insecticides or not. Unfortunately, manual counting of these insects from trap images is slow, expensive, and sometimes error-prone. Thus, developing a system, which can achieve a completely automated detection, and that can recognize and count insects is very advantageous. 

In the field of detection/segmentation of objects, many computer vision techniques have been introduced.
These techniques are, to name a few, edge detection~\cite{deriche1987}, snake contour detection~\cite{kass1988},  clustering with k-means~\cite{wagstaff2001constrained} and mean-shift~\cite{comaniciu2002}.  
For recognition/classification, usual and well-known techniques concern: Support Vector Machine~\cite{cortes1995support}, and more recently Deep Learning~\cite{lecun1995convolutional}, sometimes based on well known descriptor, such as Histogram Of Gradient, HOG~\cite{dalal2005histograms} or 
Scale Invariant Feature Transform (SIFT) descriptor~\cite{lowe2004}. 

\begin{figure}[!t]
\centering
\begin{tabular}{m{0.05\textwidth}m{0.4\textwidth}}
(a)&\includegraphics[width=0.35\textwidth]{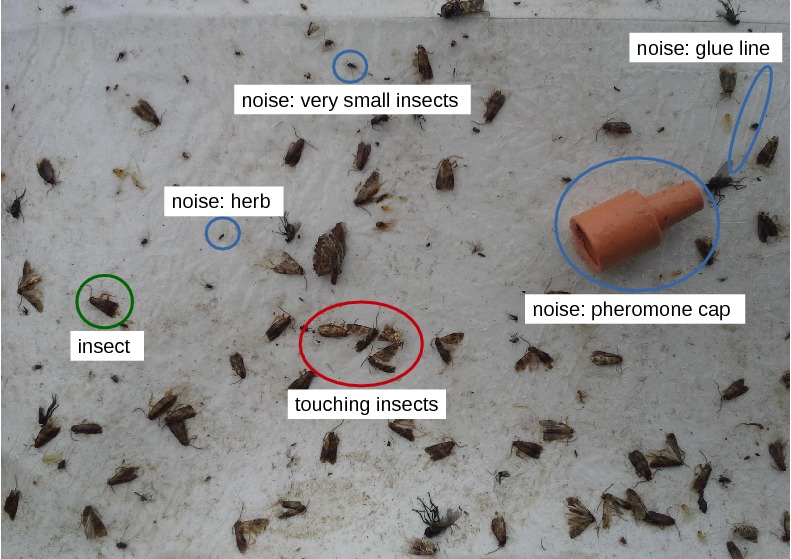}\\
(b)&\includegraphics[width=0.35\textwidth]{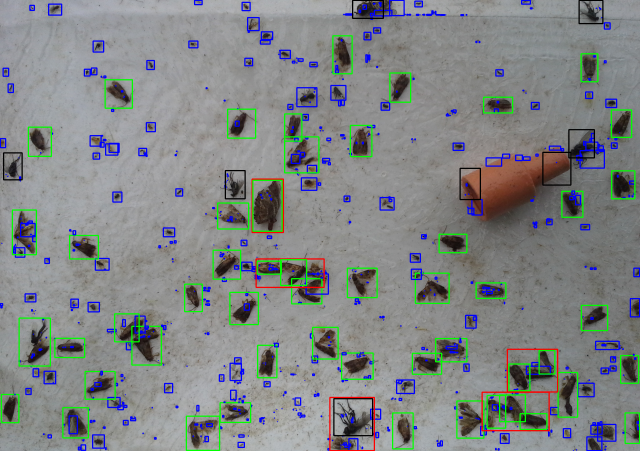}\\
\end{tabular}
\caption{An example of (a) an input image and the elements to classify, and then, (b) the classification result obtained. (a) contains the insects of interest, i.e., Lobesia Botrana, \textit{Eudemis}, a European wine moth (inside the green circle). Unfortunately, it also contains some difficulties that we generally called noise, i.e., pheromone cap, herbs and small insects (blue circles). Finally, some of the insects of interest are too close to be separated (red circle), i.e. they are touching, and this problem has to be taken into account. (b) contains the detected noise (blue rectangle), possible touching insects (red rectangle), \textit{Eudemis} (green rectangle) and other insects (black rectangle).}
\label{Trap_image}
\end{figure}

Insect classification is a challenge because it needs to recognize a small object with poor color and shape characteristics in a non-homogeneous background that can contain some difficulties. In particular, trap images may contain noise: very small insects or herbs, the pheromone cap or some lines of glue, see Fig.~\ref{Trap_image}. Moreover, since the trap is installed in outdoor environment, the trap images are also affected by illumination changes~\cite{vaib1}. Finally, touching and overlapping insects can also be found in the trap yielding a complex counting task. 

For insect recognition/classification, 
many methods have been applied to butterfly classification, 
like~\cite{kang2014identification,kaya2014application} but few publications are dedicated to agricultural insects. 
We can consider two different possibilities for doing this task. 
On one side, methods have considered insect specimens \cite{kang2014identification,wang2012new} where images contain few difficulties and are at high resolution. In this case, classification can be applied directly without the need of a segmentation step~\cite{DBLP:journals/corr/DingT16}. 
On the other side, in wild trap images, two steps are needed: a first insect segmentation before applying a classification approach in a second part. These methods encounter many challenges: low image quality, illumination, movement of the trap, movement of the moth, camera out of focus and presence of noise (such as leaves, broken wings, etc). 
In this work, we are interested in working with trap images. In the literature, for the segmentation, the authors can used color, shape and texture features~\cite{yalcin2015vision,qing2012insect}, active contour segmentation~\cite{yalcin2015vision} or morphological-based segmentation~\cite{wen2015pose}. 
All these methods do not consider the presence of touching insects in the trap. Most of descriptors used are not robust to illumination changes and do not detect occluded contours. So, this is why, in our previous work~\cite{bakkay2017automatic}, we both propose to use features that use most of the characteristics of the moth and to  take into account the problem of touching insects.

For the classification, many techniques have been tested, like $k$-nearest neighbor classification~\cite{yalcin2015vision} or 
Support vector machine, SVM~\cite{qing2012insect}. 
In~\cite{wen2015pose}, the authors even consider a pose estimation-dependent classification using deep learning. 
Deep learning based methods need large training datasets that are not always available for agricultural insects. And, in particular, in this work, we do not have such a database. 

This work aims to study the invasion of a particular moth, which is Lobesia Botrana (\textit{Eudemis}), a European vine moth, for adapting the pesticide treatment of grape culture. More precisely, wine producers usually capture this particular moth, they count the number of insects and then they analyze the evolution of this counting in order to confirm the use of the pesticides or not. Consequently, in this paper, we propose an automatic algorithm for moth classification in trap images. First, a hybrid segmentation approach, introduced in our previous ork~\cite{bakkay2017automatic}, is used to eliminate noise and to separate touching insects. 
Then, a SVM classifier is trained with a proposed multi-scale descriptor, named Histogram Of Curviness Saliency (HCS)~\cite{7532314}. This descriptor is robust to illumination changes and is able to describe the external and the internal contours of the target insect in multi-scale. Moreover, the proposed classification method can be trained with a small set of images.

In the next section, the proposed method for insect recognition is presented. To demonstrate its effectiveness, experimental results with a comparative study are detailed in Section~\ref{sec:experimentation} before the conclusion and perspectives, in Section~\ref{sec:conclusion}. 

\section{SVM Recognition based on HCS description}
\label{sec:method}

\subsection{Overview} 

\begin{figure}[!ht]
\centering{\includegraphics[width=3.3in]{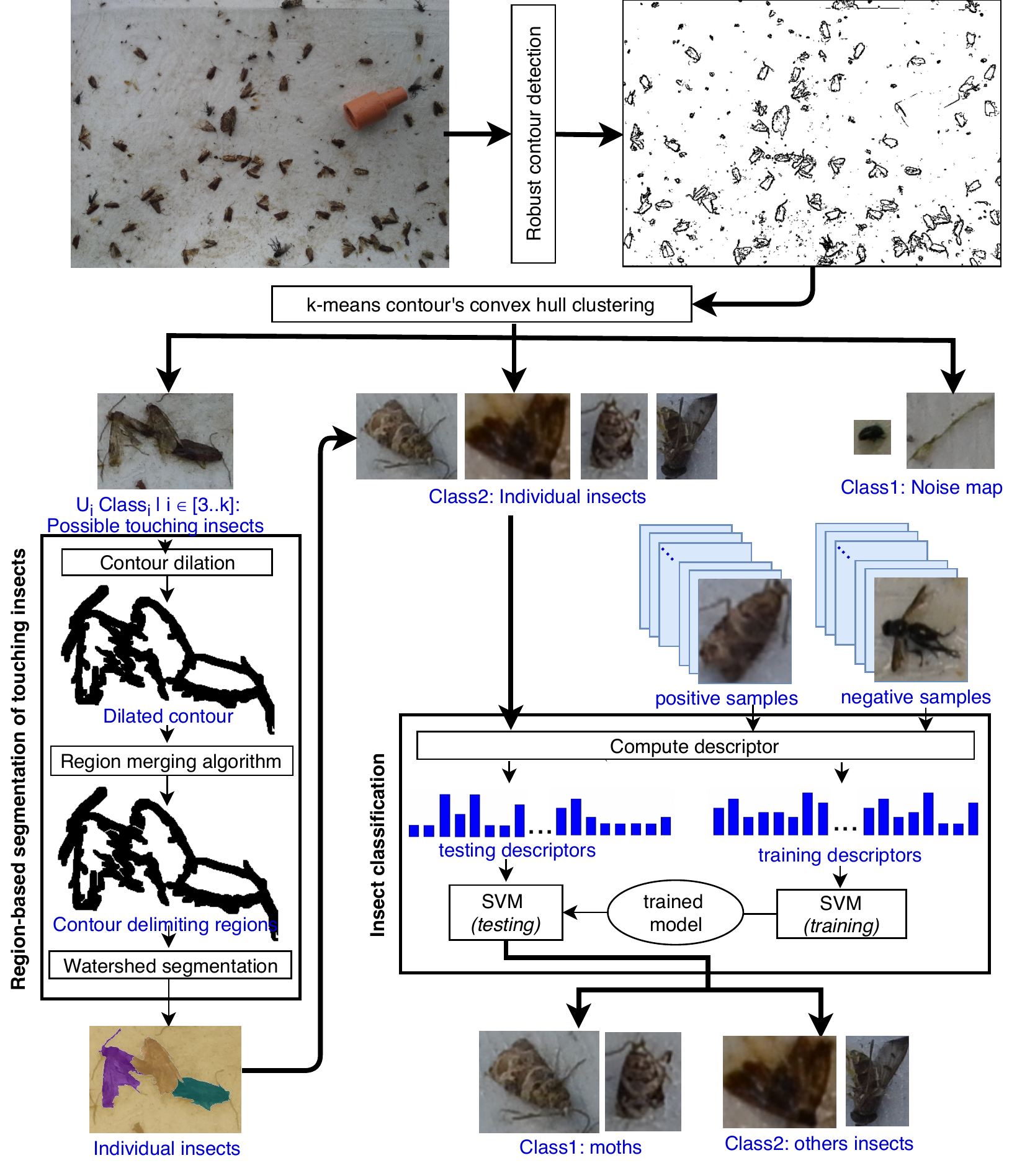}}
\caption{Overview of the proposed approach. First, a robust contour detection is applied to detect the different contours in the input image. Then, a k-means algorithm is applied to classify the estimated contours into different categories: contours due to noise (class 1), contours related to individual insects (class 2) and contours that contain touching insects (class 3 to $k$). Then, the obtained touching insects are separated by applying a region-based segmentation that contains three ordered steps: the contour dilation, the region merging algorithm and the watershed segmentation. Finally, and this is the contribution of this paper, the individual insects are classified into two categories, moths (class 1) and other insects (class 2), by using an approach based on Support Vector Machine (SVM) and the introduction of a new descriptor based on Histogram Of Gradient (HOG).}
\label{flowchart}
\end{figure}

The method allows to recognize individual moths in images of a trap that contain many difficulties (noise, touching moths and elements that are not insects), as presented in Fig.~\ref{Trap_image}. The method that starts by detecting any kind of insects, steps (1) and (2), that have been previously published and described in~\cite{bakkay2017automatic} and  then classifies the insects of interest, Lobesia Botrana, \textit{Eudemis}, from the other insects, like, for example, flies or spiders, see Fig.~\ref{flowchart}. This last step~(3) corresponds to the main contribution presented in this paper. However, for the interest of the reader, we briefly described the previous published steps (1) and (2). 

\textit{(1) Robust contour detection~--~} 
First of all, we apply a robust contour detection~\cite{7532314}.  
Then, we apply a k-means algorithm to classify the previous estimated contours into different categories. On this step, our previous work introduces an adapted criterion: the shape of the surface included in a closed contour. In fact, the shape is a significant characteristic to separate the different types of elements in this kind of scenes, i.e., it helps to distinguish between contours due to noise (class 1), contours related to individual insects (class 2) and contours that contain touching insects (class 3 to $k$). Moreover, in comparison to the state-of-the-art methods, in this approach, the number of classes is automatically selected by using the Elbow method~\cite{tibshirani2001estimating}. 

\textit{(2) Region-based segmentation~--~}
After this automated clustering step, the next task attempts to separate the obtained possible touching insects by applying a region-based segmentation. The idea is to use the contours classified into class 3 to $k$ that delimit regions as seeds for the watershed algorithm~\cite{meyer1992color}. This region-based segmentation part contains three ordered steps: the contour dilation, the region merging and the watershed segmentation. In fact, contour dilation is used to remove discontinuities of the detected contours while the region merging step avoids the over-segmentation of the watershed algorithm.  Finally, two results are possible: 
\begin{enumerate}[label=(\alph*)]
\item The watershed algorithm detects two or more regions inside the contour, thus touching insects will be separated to two or more insects. 
\item The algorithm detects only one big region and the shape of this insect is just refined. 
\end{enumerate}

All this previous work provided an encouraging detection. More precisely, the qualitative results obtained are: $87\%$ of noises or objects that are not insects are detected as noises, $70\%$ of insects are detected as insects and $82\%$ of touching insects are detected as touching insects. Moreover, the proposed method classifies $13\%$ of noises as insects, however, no noise is detected as touching insects. Then, $13\%$ of insects are detected as noises, and $17\%$ of insects are detected as touching insects. Finally, $18\%$ of touching insects are detected as insects and no touching insects are detected as noise.

\textit{(3) Insect classification~--~}
The next and final step, that corresponds to the contribution presented in this paper, concerns the insect classification. For that purpose we introduce an approach based on Support Vector Machine (SVM) that uses the Histogram Of Curviness Saliency (HCS). In the rest of the section, we will present the two aspects: how HCS is adapted to this specific recognition and how this 
descriptor is introduced in the SVM techniques. 

\subsection{Histogram of Curviness Saliency (HCS)}

In the aforementioned detection step, we used the detector proposed in~\cite{7532314} and based on curviness saliency that is an estimation of curvature. This detector generates both the magnitude and the orientation of curvature features. Thus, we naturally expand the concept of the famous classical HOG, Histogram Of Gradient~\cite{dalal2005histograms}, widely used in the feature description literature, to work on these curviness saliency features as proposed in~\cite{rashwan2018using}. More precisely, the orientation and the magnitude of the curviness saliency are used for building a descriptor called the Histogram of Curviness Saliency (HCS). That HCS is used in a sliding window fashion in a Region Of Interest (ROI) (i.e., in this work a detected insect) to generate dense features based on binning the curvature orientation over a spatial region. The orientation and the magnitude of the curviness saliency can be computed, as described in~\cite{rashwan2018using}, with:
\begin{equation}
    CS = α ((I_{xx}-I_{yy})^2 + 4I_{xy}^2),
\label{equation:CS}
\end{equation}
where  $I_{xx}$, $I_{yy}$ and $I_{xy}$ are the second derivatives of the image. 
Using this curviness saliency in multi-scale leads to this equation, as illustrated in~\cite{7532314}:
\begin{equation}
    \overrightarrow{MCS} = MCS \overrightarrow{e_{1}}.
\label{equation:MCS}
\end{equation}
It means that the multi-scale of $CS$ (MCS) of a ROI is multiplied by the eigenvector $e_{1}$ corresponding to the curviness saliency of a pixel. 

Using the same principle of HCS presented in~\cite{rashwan2018using}, we propose a descriptor that contains the orientation of the curvature of MCS binned into sparse per-pixel histograms.

\subsection{SVM classification based on HCS}

We used SVM~\cite{cortes1995support} to construct a hyperplane that separates the two classes of training data HCS descriptors (moths and other insects).
Consequently, to be adapted to the problem, we suppose that the two classes are not linearly separable, and, so, the function of this surface is given by: 
\begin{equation}
    f(\vec{x}) = \text{sgn}(\sum_{i=1}^{p} \alpha_{i}^{*}y_{i}k(\vec{x_{i}},\vec{x})+w_{0}), 
\label{equation:f}
\end{equation}
where $(\vec{x_{i}} , y_{i} )$ is related to the training data, $\vec{x_{i}}$ is the p-dimensional HCS descriptor, $\vec{y_{i}} \in \{-1,1\}$ is the class label, $k$ is a kernel function, $\alpha_{i}^{*}$ are optimal Lagrange multiplier and $w_{0}$ is the bias. For the kernel function, many kernels are possible, see~\cite{cortes1995support} to have an overview. In this work, the polynomial kernel function of degree 6 defined as:
\begin{equation}
    k(\vec{x_{i}},\vec{x_{j}}) = (\vec{x_{i}}.\vec{x_{j}}+1)^6, 
\label{equation:k}
\end{equation}
provided the best classification rate.

\section{Experimentation}
\label{sec:experimentation}

\subsection{Dataset}

We have collected annotated image patches (2865 negative patches and 746 positive patches) from moth trap images. In these images, captured under different illumination conditions, there are many insects of varying types and sizes, different elements that can induce false detections (herbs, very small insects, lines of glue, pheromone cap, etc.), see Fig.~\ref{fig_result}. 

We performed data augmentation to increase the number of positive patches from 746 to 3577. This allows also to incorporate invariance to basic geometric transformations and to noise into the classifier. Therefore, the data augmentation that we used consists in applying geometric transformations (rotation and translation), blurring, adding Gaussian noise and aspect ratio transformations to the original patches ~\cite{DBLP:journals/corr/DingT16}. 

To select training and testing sets from collected patches, we used a 5-fold cross-validation that avoids over-learning and under-learning. Patches are randomly partitioned into 5 subsamples. The cross-validation process is then repeated 5 times. In each iteration, one of the 5 subsamples is retained as the validation data for testing the model, and the remaining 4 subsamples are used as training data. This is the standard algorithm for validating any classification approach~\cite{kohavi1995study}.

\subsection{Parameter study}\label{sec:parameter}
In the proposed approach, the parameters for computing HSC descriptor are: 
\begin{enumerate}
\item detection window
\item block size
\item block stride
\item cell size
\item number of bins per cell
\end{enumerate}
These parameters are explained in details in~\cite{7532314} and~\cite{rashwan2018using}. The parameter for SVM is the soft margin $C$. 
In these experiments, detection window are set to $(64 \times 48)$, and block size to $(8 \times 8)$. In addition, block stride are set to $(4 \times 4)$ and cell size to $(4 \times 4)$. Number of bins per cell equals 9, like in the initial publication of HOG~\cite{dalal2005histograms}. Finally, the number of bins per the final descriptor equals 5940 bins. We set the soft margin parameter $C$ to $0.1$ to allow imperfect separation of classes with penalty multiplier $C$ for outliers. More precisely, in the experiments, $C$ has been tested for values in ]0, 0.5] 
and the best results are obtained with $0.1$. 

\subsection{Experimental Results}

\begin{figure}[!t]
\centering
\begin{tabular}{m{0.05\textwidth}m{0.3\textwidth}}
(a)&\includegraphics[width=0.28\textwidth,height=0.135\textheight]{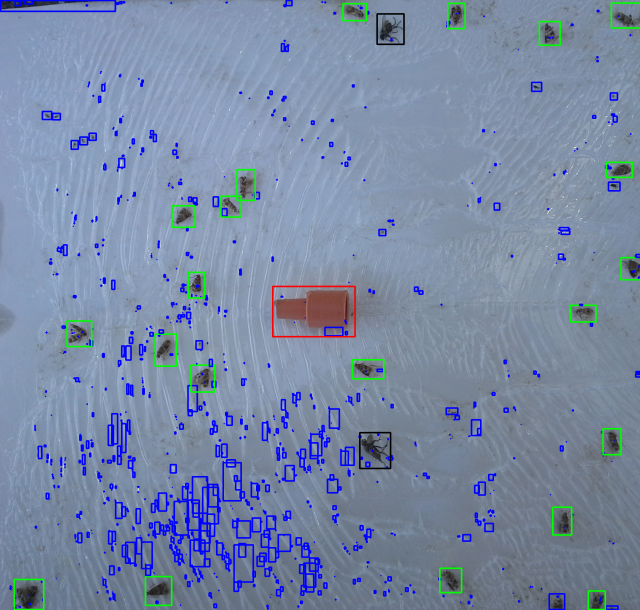}\\
(b)&\includegraphics[width=0.28\textwidth,height=0.135\textheight]{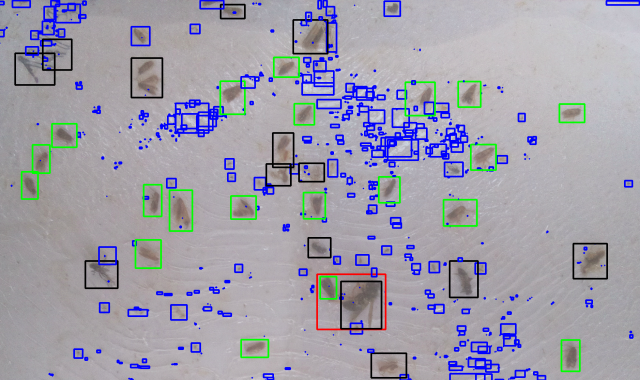}\\
\end{tabular}
\caption{Results with real trap image data set. (a) Image containing light reflections. (b) Low resolution image.}
\label{fig_result}
\end{figure}

In Fig~\ref{fig_result}, we present visual results obtained. As shown, most insects (green and purple rectangles) and noise (blue rectangles) are detected by the segmentation algorithm. In addition, touching insects (red rectangles) are separated in most of the cases. The recognition step correctly separates moths (green rectangles) to other insects (black rectangles). 

We implemented some approaches that have been presented in section~\ref{sec:introduction} and we compared them to the proposed approach on our dataset. In table~\ref{tab:results}, for all the tested methods, the accuracy rates for all tested images have been computed. The proposed method obtains the higher accuracy rate of 95\%.

\begin{table}[!htb]
\caption{Accuracies of some methods evaluated on our dataset.}
\label{tab:results}
\centering
\begin{tabular}{c|c}
\hline
\hline
\textbf{Method} &
\textbf{Accuracy(\%)} \\ 
 \hline
\cellcolor{gray!50} \textcolor{blue}{Proposed method} & 
\cellcolor{gray!50} 95.8 \% \\
\hline
Ding \textit{et al.}~\cite{DBLP:journals/corr/DingT16}& 
93.2 \% \\
\hline
\cellcolor{gray!50} Yao \textit{et al.}~\cite{qing2012insect}& 
\cellcolor{gray!50} 90.1 \% \\
\hline
Yalcin \textit{et al.}~\cite{yalcin2015vision}& 
87.6 \% \\
\hline
\hline
\end{tabular}
\end{table}

\section{Conclusion}\label{sec:conclusion}

In this paper, a classification method adapted to moth recognition has been introduced. It first uses a detection that is able to detect individual and touching insects from trap images. Second, SVM classifier is trained with a proposed multi-scale HCS descriptor which is robust to illumination changes in multi-scale way. The proposed classification method can be trained with a small set of images. Compared to state of the art methods, this new method brings the best classification rate. 
For future works, 
we plane to generalize all the proposed method to other species of insects. 



\bibliographystyle{ieeetr} 
\bibliography{db} 

\section*{Acknowledgment}
\setlength{\tabcolsep}{0.1cm}
\begin{tabular}{m{0.24\textwidth}m{0.22\textwidth}}
This work was conducted as part of the project VITI OPTIMUM 2.0. 
The project has been approved by the French competitiveness clusters: Agri Sud-Ouest Innovation and IAR. The project has been cofinanced by BPIFrance, European Regional Development Fund and the Occitania Region. 
&
\begin{center}
\includegraphics[width=0.22\textwidth]{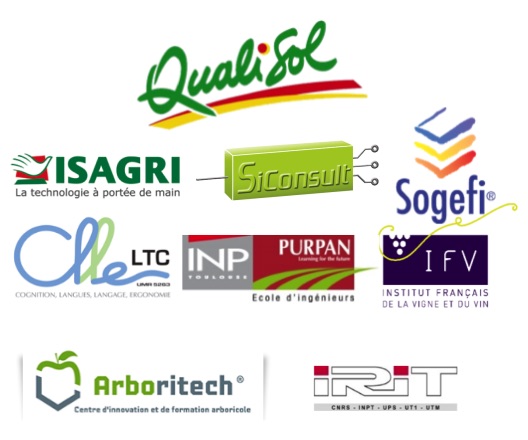} 
\includegraphics[width=0.22\textwidth]{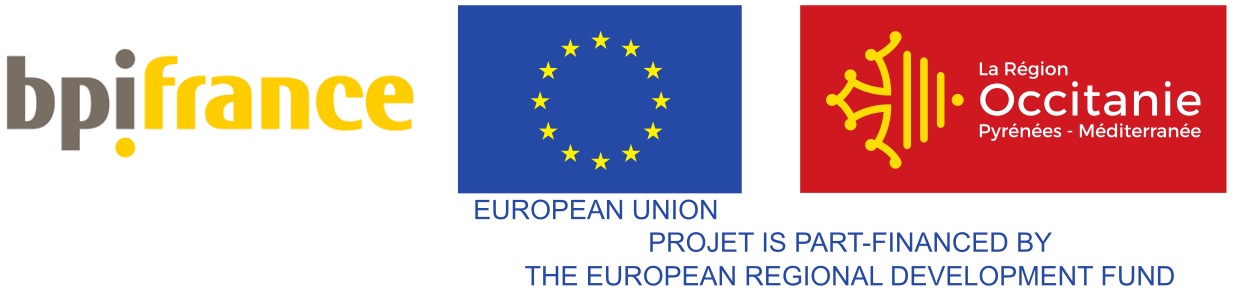}
\end{center}
\\\end{tabular}

\end{document}